\documentclass[letterpaper, 10 pt, conference]{ieeeconf}  

\IEEEoverridecommandlockouts                              

\title{\LARGE \bf
Towards Learning Object Affordance Priors from Technical Texts
}

\author{Nicholas H. Kirk$^{1}$ 
\thanks{$^{1}$ Computer Science Department, 
        Technische Universit{\"a}t M{\"u}nchen, Germany
        {\tt\small nicholas.kirk@tum.de}}
}

\usepackage{times}
\usepackage{graphicx}
\usepackage{latexsym}
\usepackage{tikz}
\usepackage[autostyle]{csquotes}

\usepackage{subcaption}
\usepackage{tikz-dependency}
\usepackage{graphicx}
\usetikzlibrary{arrows, calc, shapes,trees,positioning,arrows,chains,shapes.geometric,%
    decorations.pathreplacing,decorations.pathmorphing,shapes,%
    matrix,shapes.symbols}

\definecolor{Bittersweet}{rgb}{1.0, 0.44, 0.37}
\definecolor{BrickRed}{rgb}{0.8, 0.25, 0.33}
\definecolor{Red}{rgb}{0.8, 0.25, 0.33}
\definecolor{Brown}{rgb}{0.59, 0.29, 0.0}

\tikzset{
>=stealth',
  punktchain/.style={
    rectangle, 
    rounded corners, 
    draw=black, very thick,
    text width=10em, 
    minimum height=3em, 
    text centered, 
    on chain},
  punktelem/.style={
    rectangle, 
    rounded corners, 
    draw=red, very thick,
    text width=10em, 
    minimum height=3em, 
    text centered, 
    on chain},
  line/.style={draw, thick, <-},
  element/.style={
    tape,
    top color=white,
    bottom color=blue!50!black!60!,
    minimum width=8em,
    draw=blue!40!black!90, very thick,
    text width=10em, 
    minimum height=3.5em, 
    text centered, 
    on chain},
      punktempty/.style={
    rectangle, 
    rounded corners, 
    draw=white, very thick,
    text width=10em, 
    minimum height=3em, 
    text centered, 
    on chain},
      punktbfr/.style={
    rectangle, 
    rounded corners, 
    draw=white, very thick,
    text width=10em, 
    minimum height=1em, 
    text centered, 
    on chain},
  every join/.style={->, thick,shorten >=1pt},
  decoration={brace},
  tuborg/.style={decorate},
  tubnode/.style={midway, right=2pt},
}

\usepackage{amsmath}

\begin{document}

\maketitle
\thispagestyle{empty}
\pagestyle{empty}

\begin{abstract}

Everyday activities performed by artificial assistants can potentially be executed na{\"i}vely and dangerously given their lack of common sense knowledge.
This paper presents conceptual work towards obtaining prior knowledge on the usual modality (passive or active) of any given entity, and their affordance estimates, by extracting high-confidence ability modality semantic relations ({\em X can Y} relationship) from non-figurative texts, by analyzing co-occurrence of grammatical instances of subjects and verbs, and verbs and objects.
The discussion includes an outline of the concept, potential and limitations, and possible feature and learning framework adoption.

\end{abstract}

\section{Concept}
In the domain of autonomous robot control, artificial assistants require to know what actions can be executed on a given set of objects. Such information, defined as \textit{object affordances}, is usually obtained online by reinforcement or active learning during the execution of actions by processing percepts \cite{horton2012affordances}. However, for safe human-robot interaction, we require the robot to have, from initialization, an understanding of \textit{what actions an object can execute}, and \textit{what actions an object can be subject to}. In this scope, we claim human-written technical texts can be an informative source to construct such initial world estimate.
Such probability distribution over action-object relationships from natural language text can be performed thanks to the co-occurrence understanding of verb-noun pairs: this analysis is known in computational linguistic literature as the use of 
 distributional information of text to characterize lexical semantics, by considering statistical co-occurrence of neighboring words \cite{harris1954distributional}. 
However, the majority of current approaches make use of shallow syntactic features, of which meaningfulness is debatable for semantic characterization \cite{sahlgren2008distributional}. We therefore make use of grammatical features, for partial semantic characterization of object affordances.
While other semantic relationships employed in engineering are not easily prone to confident, automatic extraction and knowledge engineers have to recur to manual ontology insertion \cite{reeve2005survey}, the author's claim is that potentiality relationships can be robustly extracted from grammar relationships of Subject-Verb-Object (SVO) co-occurrences.
The choice of the ability modality relationship calls for the assumption that the training corpus from which we derive data has to have reliable, non-figurative subject-verb-object co-occurrence tuples. More formally, co-occurrence of every noun $s_1 \in S$ with a verb $v_1 \in V$ entails the ability of $s_1$ to perform such action $v_1$ on the co-occurring object $o_1 \in O$. In simpler terms, we assume the instance ``a robot builds a desk" implies ``\textit{a robot can build}" and ``\textit{a desk is buildable}", which the author does not consider a restrictive assumption that requires controlled authoring.

\begin{figure}[t]
\begin{subfigure}{0.25\textwidth}
\vspace{4mm}
\begin{tikzpicture}[scale=0.2, descr/.style={fill=white}]
\matrix (m) [matrix of math nodes, row sep=2.5em,
    column sep=0.5em]{
    & drawer, bottle & & \\
    & & & \\
    & door & & \vphantom{dr} \\
   \vphantom{dr} & \vphantom{dr} & arm  & \\};
 \path[->,font=\scriptsize]
 
    (m-3-2) edge [color=gray, dashed] node[descr] {\rotatebox{90}{can-contain}} (m-1-2)
    (m-3-2) edge [color=gray, dashed] node[descr] {\rotatebox{0}{can-pour}} (m-3-4)
    (m-3-2) edge [color=gray, dashed] node[descr] {\rotatebox{-90}{can-pull}} (m-4-2);
    
    \path[->,font=\scriptsize]
    (m-4-3.85) edge [color=BrickRed] (m-1-2.330)
    (m-4-3.130) edge [color=BrickRed] (m-1-2.220)
        (m-4-3) edge [color=BrickRed] (m-3-2);
\end{tikzpicture}
\end{subfigure}
\begin{subfigure}{0.15\textwidth}
\begin{tikzpicture}[scale=0.2, descr/.style={fill=white}]
  \centering
  \matrix (m) [matrix of math nodes, row sep=2.5em,
    column sep=0.5em]{
    & \vphantom{dr} & & \vphantom{dr} &\\
    &  & bottle & &\\
    & \vphantom{dr} & & arm & \\
    & &  & door, drawer & \\};
 \path[->,font=\scriptsize]
    (m-3-4) edge [color=gray, dashed] node[descr] {\rotatebox{90}{containable}} (m-1-4)
      (m-3-4) edge [color=gray, dashed] node[descr] {\rotatebox{0}{pourable}} (m-3-2)%
    (m-3-4) edge [color=gray, dashed] node[descr] {\rotatebox{-90}{pullable}} (m-4-4);
    \path[->,font=\scriptsize]
(m-2-3) edge [color=Red] (m-3-4)
(m-4-4.25) edge [color=Red] (m-3-4.325)
(m-4-4.160) edge [color=Red] (m-3-4.220);

\end{tikzpicture}
\end{subfigure}
\caption{Hypothetical representation of a dual active/passive 3-dimensional modality space
 (with predicates as dimensions) representing instances of kitchen scenario objects.}
\label{fig:meaningspace}
\end{figure}
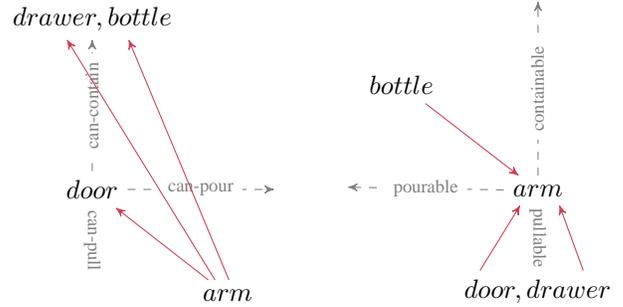

We therefore model our symbolic knowledge on potentiality as the joint probability distribution of all SVO occurrences in our training source, obtained via learning on typed dependency analysis output features of such source (see Eq. \ref{eq:jointpragma}).

\begin{equation}
\label{eq:jointpragma}
Modality(W) = P\left( S \times V \times O\right)
\end{equation}
\begin{center}
$S = \left\lbrace \forall s \in N~|~grammar\_type(s,subject)\right\rbrace \nonumber$\\
$V = \left\lbrace \forall v \in N~|~grammar\_type(v,verb)\right\rbrace \nonumber$\\
$O = \left\lbrace \forall o \in N~|~grammar\_type(o,object)\right\rbrace \nonumber$
\end{center}

From Eq. \ref{eq:jointpragma} we derive two dual joint probability distributions, which encapsulate knowledge of active and passive noun roles (Eq. 2 and Eq. 3) and can induce two distinct vector spaces, representing passive and active role information (example in Fig. \ref{fig:meaningspace}).

\begin{align}
\label{eq:dual}
Modality_{\text{active}}(W) = P(S \times V)\\
Modality_{\text{passive}}(W) = P(V \times O)
\end{align}

\section{Implementation}
In order to learn our distribution in Eq. \ref{eq:jointpragma}, 
 a possible approach is to exploit Markov Logic Networks 
(MLN)~\cite{richardson2006markov} on a set of previously extracted Stanford typed dependencies~\cite{de2008stanford}. The latter are a labeled, directed grammar relationship among pairs of words, which capture word order and relationship type (Fig. 2): when considering {\tt 'nsubj'} (subject of an action) and {\tt 'dobj'} (object of an action) labels, these can be seen as grounded action-object predicates.  
\begin{figure}[h!]
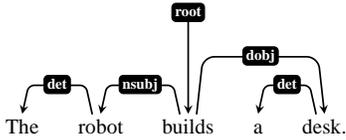

  \begin{center}
\scalebox{0.8}{

\begin{dependency}[theme=night]
"The robot builds a desk."

\begin{deptext}[column sep=.5cm, row sep=.1ex]
The \& robot \& builds \& a \& desk. \\
\end{deptext}

\deproot{3}{root}
\depedge{2}{1}{det}
\depedge{3}{2}{nsubj}
\depedge{5}{4}{det}
\depedge{3}{5}{dobj}
\end{dependency}
}

\caption{Example of words that compose a sentence instance and their typed dependencies (illustrated as labeled directed edges).}
\label{fig:wordrel}
\end{center}
\end{figure}

We can then perform learning on such grounded models thanks to MLN,
which is a knowledge representation formalism that enables probabilistic
learning and inference via the combined use of first order logic and
probabilistic undirected graphical models (i.e. \textit{Markov Random Fields}).
More formally, MLN theory defines a probability over the world $x$ as a log-linear model in 
which we have an exponentiated sum of weights $w_j$ 
of a binary feature $f_j$, and the partition function $Z$ (see Eq. \ref{eq:mln}).

\begin{equation}
\label{eq:mln}
  P\left(X = x\right) = \frac{1}{Z} ~ 
  exp\left(\sum_{j} w_j f_j \left(x\right)\right)
\end{equation}

In our case, we consider the binary formula $f_j(x)$ as an
evaluation of a logic formula representing grammar relations as predicates, and we substitute such term with $n_j(x)$, where the latter is number of true
groundings of such formula $f_j$ in $x_j$.
The MLN formalism aims to learn the stationary distribution (i.e. learn stationary weight values $w_j$) of the true groundings $n_j(x)$, possibly a sufficient heuristic condition for scalability.

\section{Discussion}
\paragraph{Related Work}

Systems which focus on the initialization parameters from ontologies (i.e. aggregates of semantic relationships and entities) do not debate how such source was populated \cite{hidayat2008learning}. Some previous literature does value mappings between language constructs and affordances, but analyze the opposite problem \cite{yuruten2012learning}.
Closer work which adopts MLN and grammar features has been proven successful for mining natural language instructions for the robotics domain \cite{nyga2012everything}, but does not focus on affordance understanding and concentrates on inferring likely action roles, while other literature does make use of MLN, but does not employ grammar feature analysis \cite{beltagy2013montague}.
Closer work does consider typed dependency extraction for semantic characterization, but does not focus on SVO tuple analysis \cite{pado2007dependency,boellasemantic}.

\paragraph{Evaluation} As the system can process a high number of noun-action relationships, we require an equally well populated ontology representing ground truth references. For activity and passivity labels, the scope might require manual annotation.

\paragraph{Potential}
Other than fulfilling the requirement of providing an initial affordance world estimate, it can provide understanding of hidden or partially observable affordances \cite{gaver1991technology}, particularly useful when objects are not in full reach of the perception array.
The vector space induction enables the use high-dimensional tensor computations for semantic characterization adopted in linguistics (such as compositionality and retrieval of neighboring entries \cite{mikolov2013efficient}), to a yet unknown extent of effectiveness within the context.
The accuracy of the learnt ontology entries can be refined via reinforcement learning during the course of task executions, or be verified thanks to an active learning process. More precisely, the SVO entries can be encapsulated in questions in a process called language generation, in order to ask for their correctness \cite{kirk2014controlled}.

\paragraph{Limitations}
Although we assume the text is confined to a technical domain, the authors of the source might make use of partly figurative wordings. 
As a result, the word frequency distribution would present bias or outliers (i.e. presence of erroneous co-occurrences of analyzed nouns or figurative nouns unrelated to known entities). Furthermore, also the independent word frequency of occurrence does not provide information regarding entity existence, and would require a form of normalization.

\paragraph{Conclusions} The linguistic and computational obstacles towards model effectiveness are manifold, and surely require the development of processes such as bias removal and outlier detection. However, this concept paper highlights the usage of technical text mining for affordance acquisition, and mainly points to the potential of induced vector spaces for retrieving objects with similar affordance, or the affordance of aggregates, and above all its practical use as initial world affordance estimate.   





\bibliography{ieeebib}
\bibliographystyle{IEEEtran}

\end{document}